\documentclass{article}

\usepackage{PRIMEarxiv}

\usepackage[utf8]{inputenc} 
\usepackage[T1]{fontenc}    
\usepackage{hyperref}       
\usepackage{url}            
\usepackage{booktabs}       
\usepackage{amsfonts}       
\usepackage{nicefrac}       
\usepackage{microtype}      
\usepackage{lipsum}
\usepackage{tablefootnote}
\usepackage{color, colortbl}
\usepackage[misc]{ifsym} 
\usepackage{soul}
\usepackage{amsmath}
\usepackage{fixltx2e}
\usepackage{afterpage}
\usepackage{caption}
\usepackage{fancyhdr}       
\usepackage{graphicx}       
\graphicspath{{media/}}     
\usepackage[sorting=none, style=numeric-comp]{biblatex} 
\definecolor{brown}{RGB}{229,225,224}
\usepackage{float}
\usepackage{mathrsfs,amsmath}
\usepackage{amsmath,amssymb,amsfonts}
\usepackage{algorithmic}

\usepackage{graphicx}
\usepackage[flushleft]{threeparttable}

\title{Neighbor2Inverse: Self-Supervised Denoising for Low-Dose Region-of-Interest Phase Contrast CT

}

\author{
Johannes B. Thalhammer$^{1-4}$ \Letter, 
Lorenzo D'Amico$^{*5}$, 
Lucy Costello$^{5}$, 
Sebastian Peterhansl$^{1,2}$, \\
\textbf{Daniel Frey$^{1,2}$, 
  Tina Dorosti$^{1-3}$, 
  Florian Schaff$^{1,2}$,
  Jannis Ahlers$^{5}$, 
  Ronan Smith$^{6}$,} \\
  \textbf{Marcus Kitchen$^{5}$,
  Franz Pfeiffer$^{1-4}$, 
  Martin Donnelley$^{6}$, 
  Daniela Pfeiffer$^{3,4}$, 
  Kaye S. Morgan$^{5}$}\\\\
  1 Chair of Biomedical Physics, Department of Physics, School of Natural Sciences, TU Munich \\
  2 Munich Institute of Biomedical Engineering, TU Munich \\
  3 Institute for Diagnostic and Interventional Radiology, TUM Klinikum \\
  4 Institute for Advanced Study, TU Munich, 85748 Garching, Germany \\
  5 X-ray Imaging Group, School of Physics and Astronomy, Monash University, Australia \\
  6 School of Medicine and Robinson Research Institute, Adelaide University, \\and Respiratory and Sleep Medicine, Women's and Children's Hospital, Adelaide, Australia \\\\
  \Letter \hspace{0.3em} \texttt{johannes.thalhammer@tum.de}
}

\begin{document}

\maketitle

\begin{abstract}
Propagation-based X-ray phase-contrast imaging (PBI) enables high-contrast visualization of lung structures and holds strong medical potential. However, safe translation to the clinic will require a substantial radiation dose reduction, which inevitably increases image noise. Supervised convolutional-neural-network-based denoising can restore image quality but depends on paired low- and high-dose datasets, which are rarely available in practice. Self-supervised methods avoid this limitation, yet most are not well adapted to the inverse problem of PBI computed tomography (CT).
We introduce \textit{Neighbor2Inverse}, a self-supervised denoising framework designed for low-dose PBI-CT that generalizes to clinical CT. Building on the Neighbor2Neighbor principle, each noisy projection is subsampled into two variants that preserve structural information but contain independent noise realizations. These are reconstructed separately, and the resulting pairs are used to train a denoising network directly in the image domain. We benchmark the proposed method against established analytical and self-supervised denoising approaches.
In region-of-interest PBI CT experiments, \textit{Neighbor2Inverse} achieves superior noise suppression while preserving fine structural details, as demonstrated by improved contrast-to-noise ratio, spatial resolution, and composite image quality metrics. Competitive performance is also observed on clinical CT data under simulated low-dose conditions.

This work has been submitted to the IEEE for possible publication. Copyright may be transferred without notice, after which this version may no longer be accessible.

Code, data, and interactive figures are available at \url{https://github.com/J-3TO/Neighbor2Inverse}.
\end{abstract}

\section{Introduction}
\label{sec:introduction}
Propagation-based X-ray phase contrast imaging (PBI) is a technique that exploits the self-interference of a coherent wavefield during free-space propagation to retrieve phase information from an  \cite{cloetens}\cite{snigirev}\cite{Paganin2002SimultaneousObject}. Unlike conventional attenuation-based imaging, PBI enhances soft-tissue contrast, making it particularly promising for applications such as cancer detection and characterization in lung imaging \cite{Yagi}\cite{PaperLucy}\cite{Ahlers2025}\cite{DAmico2025}. 
Such characterization currently often relies on invasive biopsy procedures for histological validation \cite{DeMargerie-Mellon2016Image-guidedHow}\cite{Melzer2023TumorReport}\cite{Zhang2020BiopsyStates}. If sufficient diagnostic detail can be obtained, PBI may reduce the need for biopsy or provide complementary information to improve treatment planning and understanding of tumor heterogeneity.
The minimization of radiation dose of PBI is essential for patient safety. This can be achieved by restricting the illuminated field to a region of interest (ROI), reducing the number of projections, or shortening the exposure time. The latter two, however, increase noise and introduce artifacts in the reconstructed images.

In recent years, deep learning has achieved remarkable success in denoising and artifact reduction for low-dose CT. However, acquiring large paired high-dose low-dose datasets for supervised learning is often impractical. While simulated data can help, it often fails to fully replicate the complex noise characteristics observed in real-world acquisitions, potentially introducing bias \cite{Guo2019TowardPhotographs}.
To address this, self-supervised learning methods have been proposed. The seminal Noise2Noise framework showed that, given multiple noisy observations of the same object, one can train a model by minimizing the difference between noisy pairs, without requiring a noise-free target \cite{Lehtinen2018Noise2Noise:Data}. However, this approach assumes access to multiple noisy realizations of the same image, which is not always feasible.
To overcome this limitation, several techniques have emerged. Neighbor2Neighbor generates pseudo pairs by downsampling a single image to produce two versions with statistically independent noise but nearly identical underlying signal \cite{Huang2021Neighbor2Neighbor:Images}. Similarly, ZeroShotN2N \cite{Mansour2023Zero-ShotData} utilizes checkerboard subsampling to fit a light-weight network to a single noisy image. 
Noise2Inverse operates at the projection level, dividing raw sinogram data into subsets to generate multiple sparse-view reconstructions, which are then used for self-supervised training \cite{Hendriksen2020Noise2Inverse:Tomography}. 
Proj2Proj \cite{Unal2024Proj2Proj:Reconstruction} applies perturbations to the sinogram to generate image pairs in the image domain.
Blind2Unblind \cite{Wang2022Blind2Unblind:Spots} enhances the blind-spot paradigm, which learns to predict masked pixels, by incorporating a global-aware mask mapper together with a re-visible loss to improve denoising performance. More recently, Filter2Noise \cite{sun2026filter2noiseframeworkinterpretablezeroshot} introduced dual-attention modules to predict adaptive bilateral filter parameters, SDCNN \cite{Liu2025SDCNN:Denoising} focuses on learning to disentangle signal-dependent and signal-independent noise components, and Pixel2Pixel \cite{Ma2025Pixel2Pixel:Denoising} leverages pixel-wise random sampling to generate pseudo instances.

None of these methods, however, is tailored to ROI PBI-CT, which presents a distinct set of challenges: spatially correlated image-domain noise, ROI truncation artifacts, high-pixel count images that impose significant computational demands, and the presence of phase retrieval in the reconstruction pipeline. 
Hence we propose Neighbor2Inverse to address these challenges directly. By subsampling measured projections in the spirit of Neighbor2Neighbor, we generate input pairs with near-identical object structure but independent noise, which are phase-retrieved, reconstructed, and used to train a denoising network in the image domain. 
This approach integrates the inverse nature of the denoising task with the statistical robustness of Neighbor2Neighbor without requiring paired high-dose reference data.
We investigate subsampling strategies in the 3D projection volume, evaluate data-fidelity regularization in the ROI tomographic setting, and benchmark Neighbor2Inverse against a broad range of analytical and self-supervised methods. We anticipate that both the proposed framework and the systematic comparative evaluation will serve as a valuable reference and contribute to bringing PBI closer to clinical application.

\section{Methods}
\label{sec:methods}
\subsection{Datasets}
\label{sec:Data}

Propagation-based phase contrast images were acquired at the Imaging and Medical Beamline (IMBL) of the Australian Synchrotron. The dataset consists of an inflated calf lung scanned at a 5 m propagation distance using monochromatic X-rays at 70 keV, following parameters established in prior work \cite{PaperLucy}. Agarose was injected to emulate the image signal of tumor tissue; full details of sample preparation are given in \cite{DAmico2025}. Images were captured with a PCO.edge 5.5 sCMOS detector coupled to a 25 µm Gadox phosphor scintillator and a Nikon lens, yielding an effective pixel size of 9 µm and image dimensions of 2150×2560 pixels \cite{Hall2013DetectorsSynchrotron}.
We measured one lung at six different locations with off-center acquisition at 360$^{\circ}$ and 3600 projections per measurement. At each position, scans with seven exposure times were acquired (15ms, 25ms, 33ms, 50ms, 67ms, 100ms, 200 ms). Four positions were used for training, one for validation, and one for testing. Due to sample movement and motor imprecision, measurements at different exposure times within the same position exhibit small spatial misalignments.

For clinical evaluation, 200 conventional chest CT exams were randomly drawn from the RSNA Pulmonary Embolism Detection Challenge dataset \cite{Colak2021TheDataset}, split into train/validation/test sets consisting of 120/40/40 exams. Low-dose acquisitions were simulated by forward-projecting reconstructed volumes into fan-beam sinograms with 2048 views using TorchRadon \cite{torch_radon}, then adding mixed Poisson–Gaussian noise:

\begin{equation}
    s_{\text{noisy}} = -\log\!\left(\frac{1}{\alpha}\,\mathcal{P}(\alpha \exp(-s)) + \mathcal{N}(0,\,\sigma_G^2)\right),
\end{equation}
where $s$ is the normalized sinogram, $\alpha=100,000$ is the incident photon count, and $\sigma_G=5\times10^{-4}$ is the Gaussian standard deviation, which were heuristically chosen. The noisy sinograms were re-normalized and reconstructed via Filtered Backprojection (FBP).

\subsection{Data processing, Thickness Retrieval, and Reconstruction}
\label{sec:processing}

Raw projections were first corrected for detector inhomogeneities using flat-field and dark-current correction. Dead and hot pixels were interpolated. To expand the field-of-view, projections with angles $\phi > 180^\circ$ were flipped horizontally and stitched with the corresponding projection at $\phi - 180^\circ$, effectively leading to 1,800 projections over $180^\circ$. Preprocessing steps follow the workflow in \cite{Brun2017SYRMEPWorkflows}.
Ring artifacts were mitigated using the sorting-based sinogram correction proposed by \cite{Vo2018SuperiorMicro-tomography}. The corrected and stitched projection $\phi$ is denoted as $\mathbf{p}_{\phi}(a, b)$, where $(a, b)$ represent the detector's row and column coordinates.
Projected thickness maps $\mathbf{t}_{\phi}(a, b)$ were computed using the single-material phase retrieval algorithm by Paganin et al. \cite{Paganin2002SimultaneousObject}:

\begin{equation}
\begin{split}    
\mathbf{t}_{\phi}(a, b) & = -\frac{1}{\mu}\log_e\left( \mathscr{F}^{-1} \left\{\frac{\mathscr{F}\left\{ \mathbf{p}_{\phi}(a, b) \right\}}{1 + z \delta \mu^{-1}(u^{2} + v^{2})} \right\}\right) \\
& = \mathcal{T}\mathbf{p}_{\phi}
\end{split}
\end{equation}
where $u$ and $v$ are the Fourier domain coordinates corresponding to $a$ and $b$, $z$ is the sample-to-detector distance, $\mu$ is the linear attenuation coefficient, and $\delta$ the refractive index decrement.
From the full set of projection angles, a sinogram for a given detector row $b$ is extracted as $\mathbf{t}_{b}(a, \phi)$. To mitigate truncation artifacts and bowl effects during reconstruction \cite{Arcadu2017FastTomography}, each sinogram was symmetrically padded to twice its original width by repeating the outer detector row. Subsequently, reconstruction of image slice $\mathbf{r}_b$ was performed using FBP, implemented via TorchRadon \cite{torch_radon}:

\begin{equation}
\begin{split}
\mathbf{r}_b(x, y) & = \int_0^{\pi}\int_{-\infty}^{\infty}(\mathscr{F}\left\{\mathbf{t}_b(a, \phi)\right\}(q)e^{2\pi iq\xi}|q|\text{d}q\text{d}\phi \\
& = \mathcal{R}\mathbf{t}_b
\end{split}
\end{equation}
with $\xi = x \cos(\phi) + y \sin(\phi)$, and $\mathcal{R}$ denoting the reconstruction operator. Final images were cropped to their original dimensions after padding.

\begin{figure*}[!h]
    \centering
    \includegraphics[width=\textwidth]{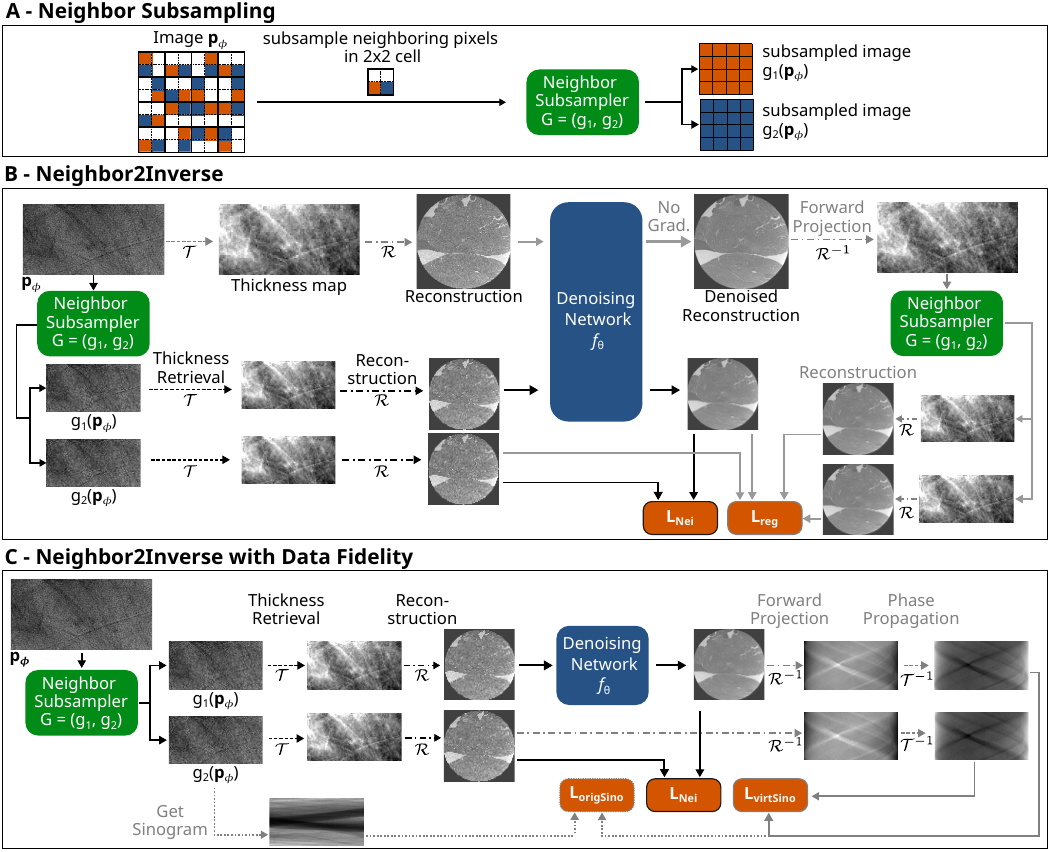}
    \caption{\textbf{A} The Neighbor subsampling algorithm. \textbf{B} The Neighbor2Inverse pipeline \textbf{C} with a data fidelity regularization instead of the regularization based on different underlying signals.}
    \label{fig:method}
\end{figure*}

\subsection{Noise Reduction Methods}
\subsubsection{Neighbor2Neighbor Revisited} \label{sec:nei2nei}
Neighbor2Neighbor ~\cite{Huang2021Neighbor2Neighbor:Images} enables self-supervised denoising by generating two sub-images from a single noisy projection.
Each noisy projection $\mathbf{p}_{\phi}$ is divided into non-overlapping $2 \times 2$ pixel patches from which two neighboring pixels are randomly selected using a subsampling operator $G$, producing two images $g_1(\mathbf{p}_{\phi})$ and $g_2(\mathbf{p}_{\phi})$ with half the spatial resolution. This is examplified in Fig. ~\ref{fig:method}A.
These sub-images share nearly identical signal content but carry uncorrelated noise, enabling the following training objective:

\begin{equation}
\label{eq:opti1}
\min_{\theta} \; \mathbb{E}_{\mathbf{p}_{\phi}} \left\| f_{\theta}(g_1(\mathbf{p}_{\phi})) - g_2(\mathbf{p}_{\phi}) \right\|^2,
\end{equation}
where $f_{\theta}$ is the denoising network with parameters $\theta$.
To compensate for small signal discrepancies introduced by subsampling, a regularization term is proposed. Under the assumption that $f_{\theta}$ is a near-perfect denoising function and that the noisy observations $\mathbf{p_{\phi}}$ are conditionally unbiased estimates of the clean signal $\mathbf{p^{*}}_{\phi}$, i.e. $\mathbb{E}_{p_{\phi} | p^{*}_{\phi}}\left\{\mathbf{p}_{\phi}\right\} = \mathbf{p^{*}}_{\phi}$ , the following holds true:

\begin{align}
\label{eq:opti2}
&\mathbb{E}_{p_{\phi} | p^{*}_{\phi}}\Big\{f_{\theta}(g_1(\mathbf{p}_{\phi})) - g_2(\mathbf{p}_{\phi}) - (g_1(f_{\theta}(\mathbf{p}_{\phi})) - g_2(f_{\theta}(\mathbf{p}_{\phi}))\big)\Big\} \notag\\
&\quad\quad\quad= 0.
\end{align}

If the gap between the underlying signals in $g_1(\mathbf{p}_{\phi})$ and $g_2(\mathbf{p}_{\phi})$ is negligible, the correction term vanishes. For non-zero signal discrepancies, the term accounts for this misalignment. Incorporating this constraint yields the full loss:

\begin{alignat}{2}
\label{eq:Nei2Nei}
L = &\left\| f_{\theta}(g_1(\mathbf{p}_{\phi})) - g_2(\mathbf{p}_{\phi}) \right\|^2 \nonumber +\\
&\gamma \left\| f_{\theta}(g_1(\mathbf{p}_{\phi})) - g_2(\mathbf{p}_{\phi}) - g_1(f_{\theta}(\mathbf{p}_{\phi})) + g_2(f_{\theta}(\mathbf{p}_{\phi})) \right\|^2 \nonumber \\
=  &L_{\text{Nei}} + \gamma L_{\text{reg}} &
\end{alignat}
with the regularization parameter $\gamma$. For theoretical background, see \cite{Huang2021Neighbor2Neighbor:Images}.

\subsubsection{Neighbor2Inverse}

The proposed Neighbor2Inverse method (Fig.~\ref{fig:method}B) extends Neighbor2Neighbor to the tomographic inverse problem by training the denoising network directly in the image domain, where spatial correlations can be more effectively exploited. Because noise in FBP images is spatially correlated, subsampling has to be performed in the projection domain before reconstruction and thickness retrieval.

14 neighboring detector rows are randomly selected from the full projection stack, which was found to be sufficient to produce results equivalent to processing the full field of view. The corresponding projections are subsampled by $G$ into $g_{1/2}(\boldsymbol{p}_{\phi})$, thickness retrieved by $\mathcal{T}$, and the central detector row $\left[\cdot\right]_b$ from each resulting thickness map is reconstructed by FBP. This yields two slices with similar signal but uncorrelated noise. One slice is passed through the network $f_{\theta}$, and the other serves as the target. 

The correction term from the Neighbor2Neighbor approach described in Eq. \ref{eq:Nei2Nei} was adapted as follows: projections are thickness retrieved, the two middle rows ($b = \{7,8\}$) are reconstructed, and denoised using $f_{\theta}$. These are then forward projected ($\mathcal{R}^{-1}$), subsampled again using $G$, and reconstructed once more, forming the correction component. The final objective becomes:

\begin{equation}
\begin{split}
\label{eq:projSubsamplingLoss}
L =\; &\left\| f_{\theta}\left(\mathcal{R}\left[\mathcal{T}g_1(\boldsymbol{p}_{\phi})\right]_{b=4}\right) 
- \mathcal{R}\left[\mathcal{T}g_2(\boldsymbol{p}_{\phi})\right]_{b=4} \right\|^2 \\
&+ \gamma \Big\| 
f_{\theta}\left(\mathcal{R}\left[\mathcal{T}g_1(\boldsymbol{p}_{\phi})\right]_{b=4}\right) 
- \mathcal{R}[\mathcal{T}g_2(\boldsymbol{p}_{\phi})]_{b=4} \\
&\quad - \mathcal{R}g_1\left(\mathcal{R}^{-1}f_{\theta}\left(\mathcal{R}[\mathcal{T}\boldsymbol{p}_{\phi}]_{b=\{7,8\}}\right)\right) \\
&\quad + \mathcal{R}g_2\left(\mathcal{R}^{-1}f_{\theta}\left(\mathcal{R}[\mathcal{T}\boldsymbol{p}_{\phi}]_{b=\{7,8\}}\right)\right)
\Big\|^2  \\
=\; & L_{\text{Nei}} + \gamma L_{\text{reg}}
\end{split}
\end{equation}

Since the projection data forms a 3D volume defined by angle $\phi$, detector width $a$, and height $b$, neighbor subsampling can be applied either in the $a$--$b$ domain (projection subsampling) or in the $a$--$\phi$ domain (sinogram subsampling), corresponding to subsampling individual projections or sinograms, respectively. Both variants follow the loss in Eq.~\ref{eq:projSubsamplingLoss}.

\subsubsection{Neighbor2Inverse Data Fidelity}

We also evaluated the inclusion of a data fidelity term in the Neighbor2Inverse approach, replacing the original regularization term. This variation is illustrated in Fig.~\ref{fig:method}C. Two variants were tested:
In the first variant, the network output is compared to the measured, subsampled sinogram $\left[g_2(\boldsymbol{p}_{\phi})\right]_{b=4}$. The corresponding loss term is 

\begin{align}
L =\; &\left\| f_{\theta}\left(\mathcal{R}\left[\mathcal{T}g_1(\boldsymbol{p}_{\phi})\right]_{b=4}\right) - \left[\mathcal{R}\mathcal{T}g_2(\boldsymbol{p}_{\phi})\right]_{b=4} \right\|^2 \nonumber \\
&+ \gamma \left\| 
T^{-1}R^{-1}f_{\theta}\left(\mathcal{R}\left[\mathcal{T}g_1(\boldsymbol{p}_{\phi})\right]_{b=4}\right)  
- \left[g_2(\boldsymbol{p}_{\phi})\right]_{b=4} \right\|^2 \nonumber \\
=\; & L_{\text{Nei}} + \gamma L_{\text{origSino}}
\label{eq:DatFidelityOrigSino}
\end{align}
 
Here, $\mathcal{T}^{-1}$ denotes forward phase propagation using the transport of intensity equation \cite{Paganin2002SimultaneousObject}\cite{Rytov1989Radiophysics}\cite{Teague1983DETERMINISTICSOLUTION.}. To mitigate artifacts introduced by image boundaries during forward projection, we pad the denoised image $f_{\theta}(\mathcal{R}\left [\mathcal{T}g_1(\boldsymbol{p}_{\phi})\right]_{b=4})$ by 1,000 pixels in the $a$-direction and 50 pixels in the $b$-direction on each side by repeating the respective outer column/row. This results in only minimal approximation error. 
Due to the ROI nature of our measurements, the original and forward-projected sinograms inherently differ, so the loss term cannot reach zero. However, since gradient-based optimization relies on the gradient rather than the absolute value of the loss, we hypothesize that this term can still effectively regularize the denoising task. Specifically, we expect that the minimum of this term corresponds to optimal denoising within the region of interest.

Alternatively, one can also forward project the padded FBP images to form a virtual sinogram limited to the region of interest \cite{Arcadu2017FastTomography}. Inspired by this, our second variant constructs a virtual sinogram from $\mathcal{R}\left[\mathcal{T}g_2(\mathbf{p}_{\phi})\right]_{b=4}$ using forward operators: $\mathcal{T}^{-1}\mathcal{R}^{-1}\mathcal{R}\left[\mathcal{T}g_2(\mathbf{p}_{\phi})\right]_{b=4}$. This leads to:

\begin{equation}
\begin{split}
L =\; &\left\| f_{\theta}\left(\mathcal{R}\left[\mathcal{T}g_1(\boldsymbol{p}_{\phi})\right]_{b=4}\right) - \left[\mathcal{R}\mathcal{T}g_2(\boldsymbol{p}_{\phi})\right]_{b=4} \right\|^2 \\
&+ \gamma \Big\| 
T^{-1}R^{-1}f_{\theta}\left(\mathcal{R}\left[\mathcal{T}g_1(\boldsymbol{p}_{\phi}\right)\right]_{b=4})  \\
&- \mathcal{T}^{-1}\mathcal{R}^{-1}\mathcal{R}\left[\mathcal{T}g_2(\mathbf{p}_{\phi})\right]_{b=4} \Big\|^2  \\
=\; & L_{\text{Nei}} + \gamma L_{\text{virtSino}}
\label{eq:DatFidelityVirtSino}
\end{split}
\end{equation}

\subsubsection{Methods for Comparison}
We compared our method against the following variation of learning based methods with publicly available code: Noise2Inverse \cite{Hendriksen2020Noise2Inverse:Tomography}, Neighbor2Neighbor in the projection domain \cite{Huang2021Neighbor2Neighbor:Images}, Blind2Unblind \cite{Wang2022Blind2Unblind:Spots}, Proj2Proj \cite{Unal2024Proj2Proj:Reconstruction}, ZeroShotN2N \cite{Mansour2023Zero-ShotData}, Filter2Noise \cite{sun2026filter2noiseframeworkinterpretablezeroshot}, and SDCNN \cite{Liu2025SDCNN:Denoising}. All methods were retrained on our PBI and the clinical dataset, staying as close to the original available code and training settings as possible.
For the PBI data, we also evaluated a supervised approach based on synthetic noise modeling. A Wasserstein GAN with gradient penalty (WGAN-GP) \cite{Gulrajani2017ImprovedGANs} was trained to generate 15 ms-like noisy projections from clean 200 ms scans. The resulting synthetic noisy–clean pairs were passed through the thickness retrieval and reconstruction pipeline, and a U-Net was trained for supervised denoising in the image domain. We refer to this model as FakeNoiseNet. For the clinical dataset, we trained the denoising model in a supervised setting directly.

In terms of analytical methods, we compared against Gaussian filtering, bilateral filtering, and wavelet denoising, BM3D \cite{Dabov2007ImageFiltering, Makinen2020CollaborativeMatching}, and total variation denoising \cite{Chambolle2004AnApplications} (all via SciPy vers. 1.15.2 \cite{Virtanen2020SciPyPython}). For the clinical dataset, parameters were optimized for highest Structural Similarity Index (SSIM). For the PBI dataset, parameter selection based on  Contrast-to-Noise Ratio (CNR) or spatial resolution yielded unsatisfactory visual results. Parameters were therefore selected by visual inspection of denoising quality.

\subsubsection{Network Architectures}  
In the Noise2Inverse and Neighbor2Inverse approaches, as well as in the FakeNoiseNet, we used the U-Net architecuture as the backbone \cite{Ronneberger2015U-net:Segmentation}. We followed the implementation described in \cite{Huang2021Neighbor2Neighbor:Images}, which includes five max pooling layers, a leaky ReLU activation function, and transpose convolutions with a kernel size of 2x2 for the upsampling operation.

For training the Wasserstein Generative Adversarial Network with Gradient Penalty (WGAN-GP), we employed the same U-Net architecture as the generator function, while the EfficientNetV2-S served as the discriminator \cite{TanEffNet}. 

\subsubsection{Training Details}

All models were implemented in PyTorch 2.6.0 and PyTorch Lightning 2.5.0~\cite{Paszke2019PyTorch:Library, Falcon_PyTorch_Lightning_2019}, trained on an NVIDIA A100 (80~GiB VRAM), and optimized with Adam (default hyperparameters)~\cite{Kingma2015Adam:Optimization}. The initial learning rate was selected using the learning rate finder~\cite{Smith2017CyclicalNetworks} and subsequently scheduled via \texttt{ReduceLROnPlateau} with a patience of 5 epochs and a reduction factor of 0.5. Final models were selected based on the lowest validation loss.
The Noise2Inverse model was trained with 768×768 patches, a mini-batch size of 3, and L2 loss. We used the X:1 sampling strategy, where the input volume was reconstructed from 3/4 of the projection angles and the target from the remaining 1/4. Neighbor2Inverse models were trained on full images (4675×4675 pixels) with a batch size of 1. Gradients were accumulated over four batches before each parameter update. The weight parameter $\lambda$ in Eq.~\ref{eq:Nei2Nei} was linearly scheduled as $2\cdot\frac{\text{epoch}}{100}$ over 100 epochs, following \cite{Huang2021Neighbor2Neighbor:Images}.
The projection subsampling model was trained with and without $L_{\text{reg}}$ to assess its effect. The sinogram subsampling model was trained using $L_{\text{Nei}}$ only, as $L_{\text{reg}}$ did not have a substantial effect in the proj. subsampling case. For the data fidelity variants, $\gamma$ was chosen such that $L_{\text{Nei}}$ and the fidelity term contribute equally, yielding $\gamma=0.1$ for $L_{\text{origSino}}$ and $\gamma=50.0$ for $L_{\text{virtSino}}$. To assess robustness, a sensitivity analysis was performed by varying $\gamma$ by factors of 0.1 and 10 around the selected values. All models were trained until convergence, as determined by the validation loss.
Training without regularization required approximately 24~h and 13~GiB of VRAM for both subsampling variants. Adding $L_{\text{reg}}$ increased this to 46~h and 28~GiB, while $L_{\text{origSino}}$ and $L_{\text{virtSino}}$ required approximately 45~h and 60~h, and 45~GiB and 46~GiB of VRAM, respectively. Average inference time was 1.01~s per slice for all Neighbor2Inverse models, as all variants share the same U-Net architecture at inference.
To assess robustness to undersampling, all Neighbor2Inverse models were also trained on 15 ms reconstructed images using only half (900) of the original projection angles. This allowed us to evaluate the models' ability to suppress both noise and angular undersampling artifacts.
Lastly, we trained the Neighbor2Inverse model without a regularization term on all measured subsets (without test measurement), ranging from 200 ms to 15 ms exposure time. For each exposure setting, we additionally simulated sparse-view acquisitions by reconstructing from every $n$th projection, with $n \in [1, 4]$. 

\subsection{Quantitative Analysis}  
For the PBI data, image quality was assessed using three metrics: CNR, spatial resolution (SR), and a composite image quality index (\(Q\)), defined as $Q = \frac{CNR}{SR}$.  
For calculation, ten representative slices were uniformly sampled from the reconstructed test volume. CNR was estimated using four pairs of neighboring ROIs for each slice, containing either homogeneous soft tissue or air. It was computed as:  

\begin{equation}
CNR = \frac{\overline{I}_{ST} - \overline{I}_{air}}{\sigma_{ST}},
\end{equation}  

where \(\overline{I}_{ST}\) and \(\overline{I}_{air}\) denote the mean intensities of the soft-tissue and air ROIs, respectively, and \(\sigma_{ST}\) is the standard deviation of the soft-tissue ROI.  

To assess SR, four ROIs per slice were selected containing a clearly delineated tissue–air boundary. 
For each ROI, multiple intensity profiles perpendicular to the interface were extracted and averaged to obtain a representative edge profile, implemented using Fileswell  \cite{Fileswell}.
The averaged edge profile was fitted with an error function, which was subsequently differentiated to obtain a Gaussian. 
The full width at half maximum (FWHM) of the resulting Gaussian provided an estimate of the spatial resolution.  
For the clincial dataset, we assessed image quality using SSIM and Peak Signal-to-Noise Ratio (PSNR) on the test set.

\section{Results}
\label{sec:results}

\subsection{Denoising Results}

\begin{figure*}[h]
    \centering
    \includegraphics[width=\textwidth]{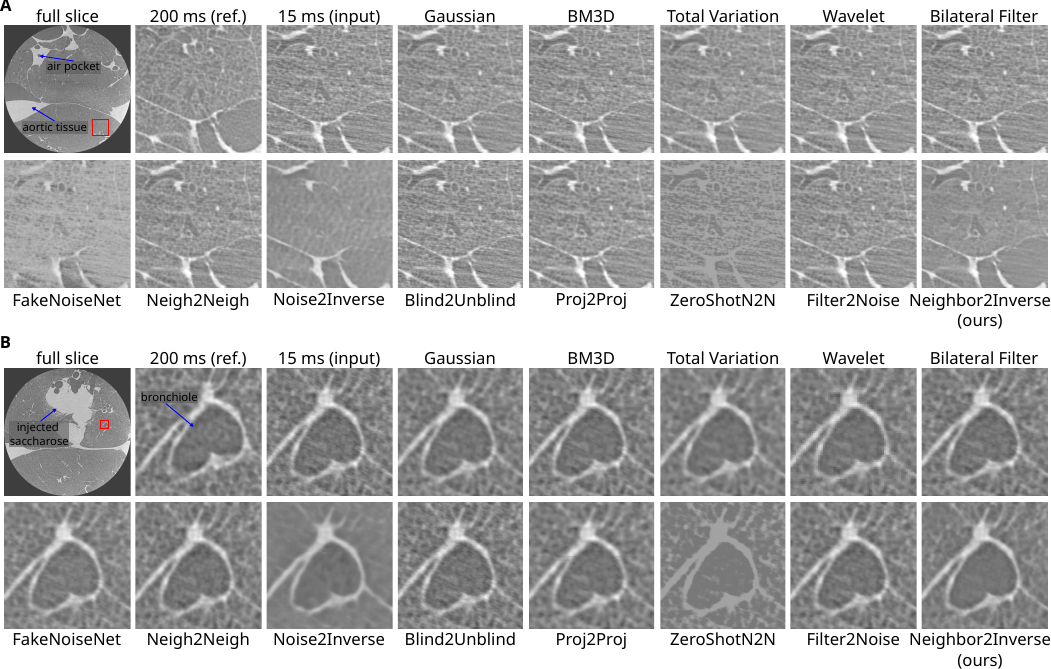}
    \caption{Comparison of denoising methods applied to two 15 ms reconstructed propagation-based CT images from the test set  with the corresponding 200 ms reference.
    The first image in \textbf{A} and \textbf{B} depict the full image (4675×4675 pixels), the subsequent images show zoomed-in regions for detailed comparison, at resolutions of 600×600 (A) and 300×300 pixels (B), respectively.}
    \label{fig:CompareNei2Inv}
\end{figure*}

\begin{figure*}[h]
    \centering
    \includegraphics[width=\textwidth]{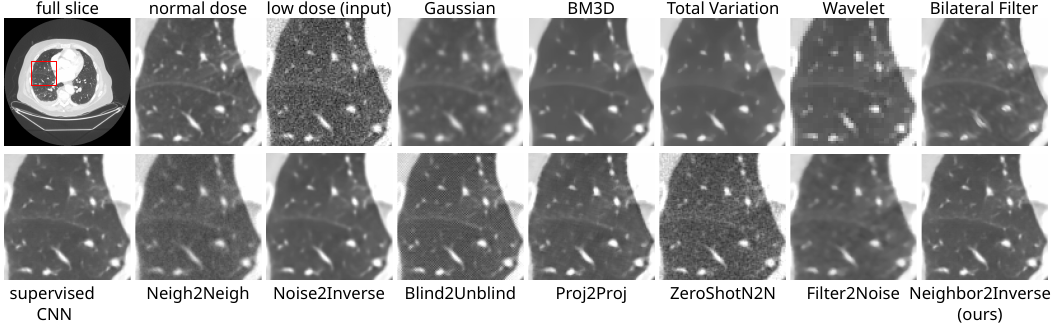}
    \caption{Comparison of denoising methods applied to clinical CT data with simulated noise added at projection level. The first image depicts the full 512x512 slice, the subsquent images show a 100x100 pixels zoom in. All images are displayed in the lung window (-1,350 HU - 150 HU).}
    \label{fig:CompareNei2Inv_PE}
\end{figure*}

\begin{figure*}[h]
    \centering
    \includegraphics[width=\textwidth]{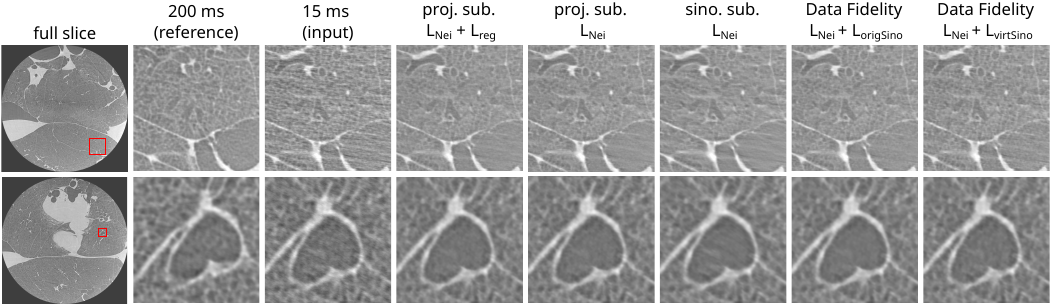}
    \caption{Denoising results of various Neighbor2Inverse approaches applied to two 15 ms reconstructed propagation-based CT image from the test set, with the corresponding 200 ms reference shown for comparison. 
    Compared are Neighbor2Inverse with projection subsampling using L$_{\text{Nei}}$ + L$_{\text{reg}}$, projection subsampling with L$_{\text{Nei}}$ only, sinogram subsampling approach with L$_{\text{Nei}}$ only, Neighbor2Inverse with L$_{\text{origSino}}$ as data fidelity term, Neighbor2Inverse with L$_{\text{virtSino}}$ as data fidelity term. 
    The first column shows the full reconstructed slices (4675×4675 pixels). The remaining images display zoomed-in regions for detailed comparison at 600×600 (upper row)and 300×300 pixels (bottom row), respectively.}
    \label{fig:VariationsOfNei2Inv}
\end{figure*}

Fig.~\ref{fig:CompareNei2Inv} shows representative 200~ms and 15~ms slices from the test set alongside denoised outputs from all compared methods. Anatomical structures including air pockets, aortic tissue, and injected saccharose are clearly visible, and individual alveoli and bronchioles can be discerned in the magnified regions. All methods reduce noise relative to the 15~ms scan, but notable differences remain.
In \ref{fig:CompareNei2Inv}A, most methods fail to suppress the horizontal stripe artifacts. FakeNoiseNet introduces substantial structural distortions, likely caused by a domain gap between the synthetic noise used during training and the actual noise in the input images.
In \ref{fig:CompareNei2Inv}B, Gaussian, TV, and wavelet filtering reduce noise but introduce oversmoothing. BM3D and bilateral filtering better preserve spatial resolution but leave residual noise. Neighbor2Neighbor, Blind2Unblind, Proj2Proj, and Filter2Noise also retain some residual noise. FakeNoiseNet performs well in this case.
For both subfigures, Noise2Inverse achieves strong noise suppression but removes fine structural details such as alveolar boundaries. ZeroShotN2N yields subpar results. Neighbor2Inverse with projection subsampling and $L_{\text{Nei}} + L_{\text{reg}}$ achieves effective noise reduction while preserving small structures, though some horizontal stripe artifacts remain visible in \ref{fig:CompareNei2Inv}A. SDCNN did not produce sensible results on either the PBI or the clinical dataset and is therefore excluded from the comparison.
In Fig. \ref{fig:CompareNei2Inv_PE}, denoising results on simulated low dose chest CTs are depicted. Again, all methods lead to a substantial noise reduction. The analytical methods, as well as Noise2Inverse, Proj2Proj, and Filter2Noise lead to an oversmooth image, Neighbor2Neighbor, Blind2Unblind, and ZeroShotN2N still leave a lot of noise in the image. Again, Neighbor2Inverse leads to an effective noise suppresion, while maintaining image sharpness, leading to visual results very similar to the results of the supervised approach.
Fig. \ref{fig:VariationsOfNei2Inv} presents denoising results of the 15 ms scan using different variations of the Neighbor2Inverse approach. Comparing the projection subsampling variant with and without the regularization term L$_{\text{reg}}$, no substantial difference in performance is observed. This suggests that the signal discrepancy between the subsampled inputs $g_1(p_{\phi})$ and $g_2(p_{\phi})$ is negligible. Given that the system’s effective resolution is influenced by factors such as source blur and scintillator scattering, rather than being limited by pixel size, this observation appears plausible. 
The network trained using sinogram subsampling leads to slightly more residual noise in comparison to the projection subsampling variant. 
The outputs of the two models trained with a data fidelity term are visually similar to the projection subsampling approach. However, the model trained with L$_\text{virtSino}$ exhibits marginally higher noise levels.

\begin{figure*}[h]
    \centering
    \includegraphics[width=\textwidth]{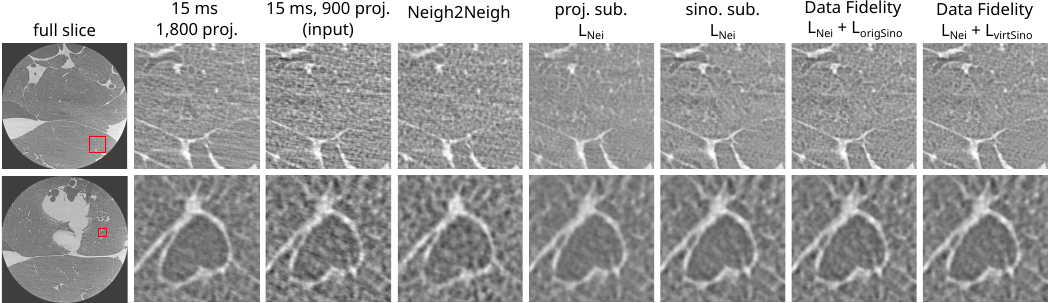}
    \caption{Denoising results of various Neighbor2Inverse and the Neighbor2Neighbor approaches applied to a sparse 15 ms propagation-based CT image from the test set reconstructed with only half (900) the measured projections, with the corresponding full-view (1800 projections) 15 ms scan shown for comparison. Compared are Neighbor2Neighbor denoising directly on the projections, Neighbor2Inverse projections subsampling L$_{\text{Nei}}$ only, Neighbor2Inverse sinogram subsampling L$_{\text{Nei}}$ only, Neighbor2Inverse with L$_{\text{origSino}}$ as data fidelity term, Neighbor2Inverse with L$_{\text{virtSino}}$ as data fidelity term. 
    The first column shows the full reconstructed slices (4675×4675 pixels). The remaining images display zoomed-in regions for detailed comparison at 600×600 (upper row)and 300×300 pixels (bottom row), respectively.}
    \label{fig:VariationsOfNei2InvSparse}
\end{figure*}

Fig. \ref{fig:VariationsOfNei2InvSparse} compares various Neighbor2Inverse variants and the Neighbor2Neighbor approach applied to sparse-view data, where only 900 of the original projection angles were used to reconstruct the 15 ms scans. This undersampling degrades image quality and introduces visible artifacts.
The Neighbor2Neighbor model reduces noise but does not address artifacts introduced by the sparse sampling.
The Neighbor2Inverse model trained with projection subsampling reduces both noise and artifacts but at the cost of attenuated contrast, particularly in fine structures.
The sinogram subsampling variant preserves more contrast, though residual noise and horizontal artifacts remain.
Neighbor2Inverse models with data fidelity terms provide better feature retention and contrast but less noise suppression, indicating that the data fidelity term enables a balance between denoising and structural preservation.

\begin{figure}[h]
    \centering
    \includegraphics[width=.6\columnwidth]{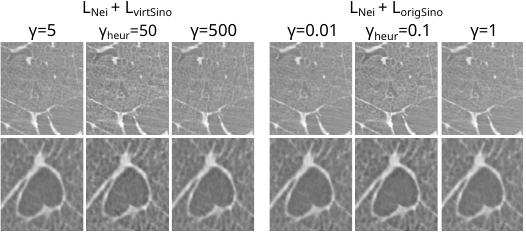}
    \caption{Sensitivity Analysis. The initial heuristically chosen weight $\gamma_{\text{heur}}$ for L$_{\text{origSino}}$ and L$_{\text{virtSino}}$ was varied by a factor 0.1 and 10, respectively, to investigate its effect on the image quality.}
    \label{fig:SensitivityAnalysis}
\end{figure}

The sensitivity analysis presented in Fig.~\ref{fig:SensitivityAnalysis} demonstrates that image quality is largely stable across the tested values of $\gamma$, indicating that the method is robust to the exact parameter choice within a reasonable range.

\begin{figure}[t]
    \centering
    \includegraphics[width=\columnwidth]{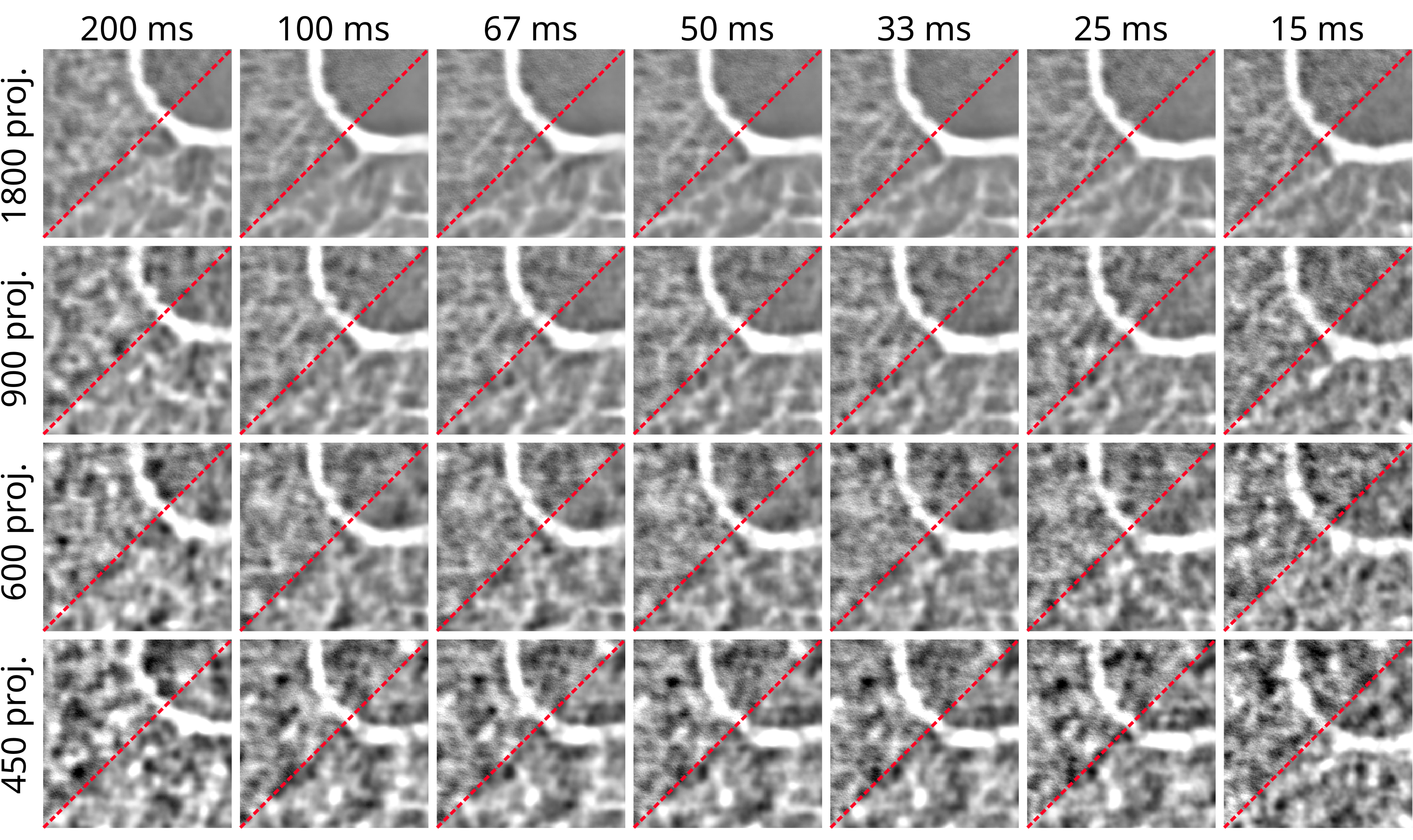}
    \caption{Denoising results of Neighbor2Inverse with varying exposure times and projection views. The displayed images are a 200x200 pixel excerpt of the originally 4675x4675 pixel image. Each image is divided diagonally; the upper triangle shows the original image, while the lower triangle shows the denoised output using the Neighbor2Inverse model (projection subsampling, only L$_{\text{Nei}}$).}
    \label{fig:GridLowExpSparse}
\end{figure}

Fig.~\ref{fig:GridLowExpSparse} compares dose reduction strategies: decreasing exposure time, reducing the number of projections, or combining both. Each image is divided diagonally, with the upper triangle showing the original image and the lower triangle the Neighbor2Inverse denoised output (proj. subsampling, $L_{\text{Nei}}$ only).
Reducing the number of projections causes more severe image degradation than reducing exposure time, likely because the thickness retrieval step suppresses noise and acts as an implicit filter. Neighbor2Inverse substantially improves image quality for low-exposure acquisitions, but does not correct undersampling artifacts in sparse-view reconstructions.

\subsection{Quantitative Evaluation}

\begin{figure*}[!h]
    \centering
    \includegraphics[width=\textwidth]{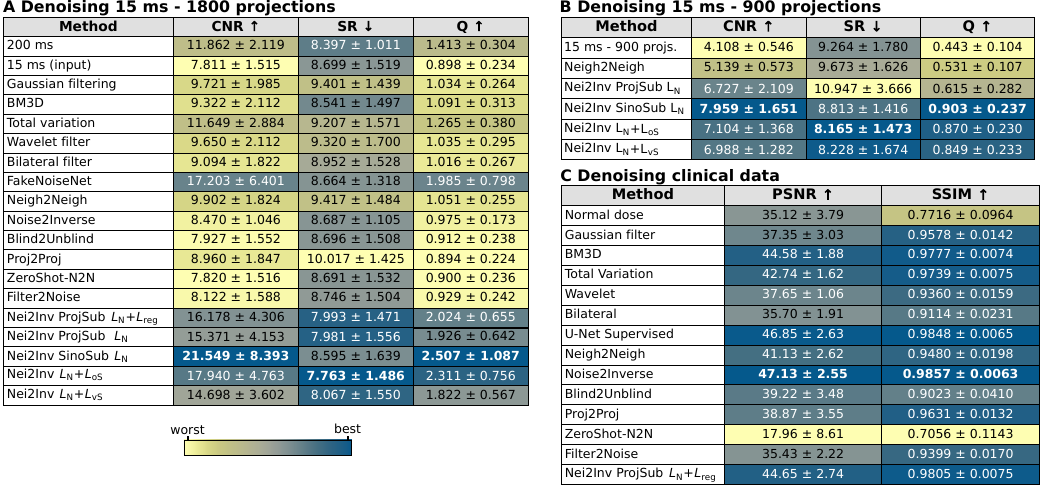}
    \caption{Quantitative evaluation of the different denoising methods applied on \textbf{A} 15ms measurements reconstructed with 1,800 projections and \textbf{B} 900 projections in terms of contrast-to-noise ratio (CNR), spatial resolution (SR) in pixels, and image quality Q=CNR/SR. \textbf{C} shows PSNR and SSIM values of the clinical dataset. Depicted are mean values along with standard deviations. Corresponding example images are shown in Fig. \ref{fig:CompareNei2Inv}-\ref{fig:VariationsOfNei2InvSparse}.}
    \label{fig:Metrics}
\end{figure*}

Fig.~\ref{fig:Metrics} summarizes the quantitative evaluation of denoising methods. Panel~A shows results for 15~ms measurements reconstructed from all available 1,800 projections. As expected, reducing the exposure time from 200~ms to 15~ms decreases CNR and increases SR. The Neighbor2Inverse method with sinogram subsampling achieves the highest overall image quality index (\(Q\)), followed by Neighbor2Inverse with the \(L_{\text{origSino}}\) data fidelity term and Neighbor2Inverse with projection subsampling plus \(L_{\text{origSino}}\). Notably, these methods even outperform the 200~ms reference in terms of the combined quality metric.  

Fig.~\ref{fig:Metrics}B reports results for 15~ms measurements reconstructed from only 900 projections. Again, the highest \(Q\) is obtained with Neighbor2Inverse and sinogram subsampling, closely followed by Neighbor2Inverse with \(L_{\text{origSino}}\).  

Interestingly, the sinogram subsampling strategy yields higher quantitative scores, despite appearing visually inferior to the projection subsampling in Fig.~\ref{fig:VariationsOfNei2Inv}. This discrepancy highlights a limitation of the ROI-based evaluation: the selected homogeneous regions emphasize noise suppression and contrast, but are insensitive to subtle distortions. A similar effect is evident for RecoFakeNoiseNet, which achieves a high CNR despite introducing clear structural artifacts, as shown in Fig.~\ref{fig:CompareNei2Inv}A.  

Fig.~\ref{fig:Metrics}C reports PSNR and SSIM on the clinical dataset, where Noise2Inverse performs best. As is well established, both metrics tend to favor over-smoothed images, and results should be interpreted with this bias in mind.

\section{Conclusion}
\label{sec:conclusion}
In this work, we introduced Neighbor2Inverse, a self-supervised denoising approach for propagation-based X-ray phase-contrast computed tomography. By training a U-Net in the image domain rather than in the projection domain, the method achieves more effective suppression of residual noise in the final images.
Compared with the established analytical and self-supervised methods, Neighbor2Inverse achieves a better balance between noise reduction and structural preservation, avoiding the excessive loss of fine details and shows robust performance, when applied to other datasets.
Our experiments show that data regularization is not essential for effective training in this setting. Incorporating data-fidelity constraints improves structural preservation at the cost of slightly higher residual noise. A more thorough investigation of data-fidelity formulations represents a promising direction for future work.
Dose-reduction studies further indicate that reducing exposure time per projection is less detrimental to image quality than reducing the number of projections. Neighbor2Inverse consistently enhances image quality in low-dose regimes and remains robust under moderate undersampling, though performance declines under severe undersampling.
Quantitative evaluation confirms improvements in contrast-to-noise ratio, spatial resolution, and a composite quality index on the PBI data, as well as in PSNR and SSIM values on the clinical dataset. However, discrepancies between numerical metrics and visual impression highlight a limitation of these metrics: They are insensitive to distortions in fine anatomical structures and systematically favor over-smoothed images. Since the clinically relevant image properties are ultimately defined by the downstream task, task-based evaluation is an important direction for future work. Nevertheless, the results presented here demonstrate that with our proposed method substantial image quality improvements are achievable.


\newpage
\printbibliography 
\end{document}